# Neural Decoding of Overt Speech from ECoG Using Vision Transformers and Contrastive Representation Learning

Mohamed Baha Ben Ticha[1], Xingchen Ran[1,2], Guillaume Saldanha[1], Gaël Le Godais[1], Philémon Roussel[1], Marc Aubert[1], Amina Fontanell[1,5], Thomas Costecalde[3], Lucas Struber[3], Serpil Karakas[3], Shaomin Zhang[2], Philippe Kahane[4], Guillaume Charvet[3], Stéphan Chabardès[3,5], Blaise Yvert[1]

Email: mohamed-baha.ben-ticha@inserm.fr and blaise.yvert@inserm.fr
[1]Univ. Grenoble Alpes, Inserm, U1216 Grenoble Institutes Neuroscience, Grenoble, France
[2]Zhejiang University, Qiushi Academy for Advanced Studies, Hangzhou, People's Republic of China
[3]Univ. Grenoble Alpes, CEA, LETI, Clinatec, Grenoble, France
[4] Univ. Grenoble Alpes, CHU Grenoble Alpes, Department of Neurology, Grenoble, France
[5]Univ. Grenoble Alpes, CHU Grenoble Alpes, Department of Neurosurgery, Grenoble, France

*Abstract*— Speech Brain Computer Interfaces (BCIs) offer promising solutions to people with severe paralysis unable to communicate. A number of recent studies have demonstrated convincing reconstruction of intelligible speech from surface electrocorticographic (ECoG) or intracortical recordings by predicting a series of phonemes or words and using downstream language models to obtain meaningful sentences. A current challenge is to reconstruct speech in a streaming mode by directly regressing cortical signals into acoustic speech. While this has been achieved recently using intracortical data, further work is needed to obtain comparable results with surface ECoG recordings. In particular, optimizing neural decoders becomes critical in this case. Here we present an offline speech decoding pipeline based on an encoder-decoder deep neural architecture, integrating Vision Transformers and contrastive learning to enhance the direct regression of speech from ECoG signals. The approach is evaluated on two datasets, one obtained with clinical subdural electrodes in an epileptic patient, and another obtained with the fully implantable WIMAGINE epidural system in a participant of a motor BCI trial. To our knowledge this presents a first attempt to decode speech from a fully implantable and wireless epidural recording system offering perspectives for long-term use.

*Keywords*— Speech decoding, ECoG, Brain-computer Interfaces, deep neural networks.

## I. INTRODUCTION

Loss of speech due to brain injury or neurodegenerative disease severely impacts the quality of life and often leads to social isolation. Brain Computer Interfaces (BCI) that decode neural activity into spoken language [1], [2], [3] offer a compelling solution to restore communication. Recent advances in this field follow a seminal study proposing a fully implantable solution based on a small set of neurotrophic electrodes chronically implanted in the ventral motor cortex that enabled the real-time decoding of vowels in real time. Yet, the limited number of neural signals that could be gathered with these electrodes called for other types of more extended neural interfaces to ensure intelligible speech reconstruction. In this respect, a number of subsequent studies showed that subdural ECoG signals were highly informative of the articulatory and auditory content of produced speech [4], [5], and that overt speech produced by epileptic patients implanted for the exploration of the origin of their seizures could be reconstructed offline with increasingly high intelligibility [6], [7], [8], [9], [10], [11], [12], [13], [14], [15]. These works paved the way towards several recent highly convincing demonstrations of real-time closed-loop speech synthesis in people with severe paralysis no longer able to speak. These proof-of-concepts of functional speech BCIs rely either on subdural ECoG recordings [16], [17], [18], [19], [20], [21], or, more recently, on intracortical recordings with Utah arrays [22], [23], [24]. These later studies based on spiking activity achieved the best decoding and speech reconstruction performance. Yet, intracortical implants being prone to degradation [25], [26] or encapsulation [27] on the long term or providing signals of variable quality from subjects to subjects [28], achieving robust speech BCIs very stable on the long term will likely require improving these technologies or considering other types of implants. Moreover, minimizing surgical invasiveness to maximize patient safety is also at stake. To this end, speech BCIs based on intracranial stereotactic EEG with depth electrodes has been considered but do not so far enable intelligible speech reconstruction [29], [30]. In this context, ECoG thus currently remains a technology of choice to build stable speech BCIs, as surface grids offer stable signals on the long term after the phase of tissue reaction post-implantation has stabilized [31], [32]. However, this technique providing signals of lower resolution than intracortical recordings, optimizing decoding algorithms for speech reconstruction from surface neural signals is at stake. In an earlier study, intelligible speech synthesis from ECoG data in a regression approach was demonstrated using an intermediate articulatory representation [11]. However, there has yet been no demonstration of intelligible speech reconstruction from ECoG signals through a direct decoding pipeline regressing acoustic coefficients of speech directly from brain signals. To this end, advanced deep-learning algorithms are at stake. Recent studies have introduced transformer architectures to reconstruct speech from neural signals [24], [33], [34]. Here, we propose to combine such approach for the reconstruction of continuous speech, with a contrastive learning (CL) scheme previously introduced for discrete classification of perceived speech items from non-invasive brain recordings [35].

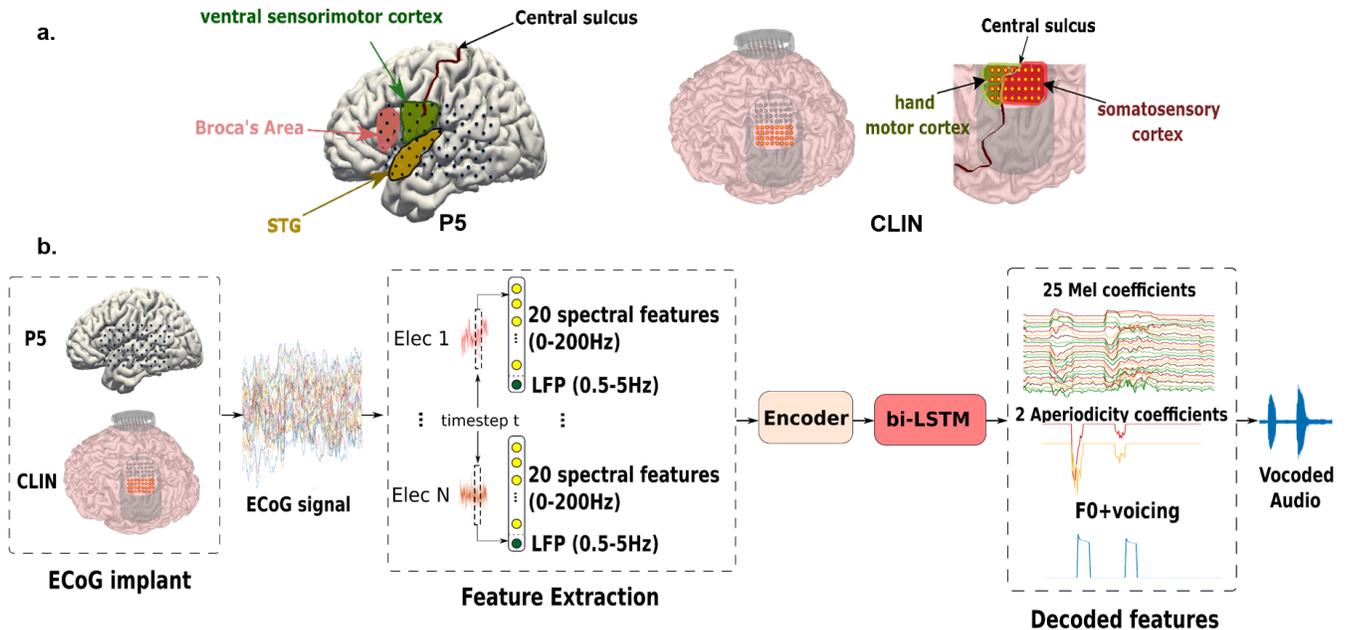

**Figure 1.** General decoding pipeline and detailed participants brain coverage. (a) presents the brain coverage for P5 (left) and CLIN (right). (b) presents the adopted global offline decoding pipeline starting with the extraction of 21 neural features per electrode signal subsequently forwarded to the encoder-decoder architecture transformed afterwards into audio waveform.

Moreover, we evaluate this paradigm on both subdural and epidural ECoG data. To this end, we introduce the usage of the WIMAGINE system [36], a chronic and fully implantable epidural brain implant previously used for motor compensation [37], [38] that does not require the opening of the dura matter and displays very stable neural signals over many years [39], [40].

## II. METHODS

### A. Participants

This study involved two participants. The first, called P5, is a female epileptic patient aged 38 who was implanted with a 9x8 subdural ECoG grid (PMT) covering a substantial portion of the left cerebral hemisphere (Figure 1.a left), as part of the Brainspeak Clinical Trial (NCT02783391). The second participant, called CLIN, is a 35-year-old male tetraplegic patient enrolled in a clinical trial aiming at controlling an exoskeleton from epidural ECoG signals (NCT02550522). He was chronically implanted with the epidural WIMAGINE system [36] over the hand motor and adjacent somatosensory cortices (Figure 1.a right). Although these areas are not traditionally associated with speech production, previous research demonstrated that the dorsal laryngeal motor cortex (dLMC) located ventrally adjacent to the hand motor cortex plays a critical role in pitch control [41]. This motivated the inclusion of CLIN in this speech decoding study. CLIN was implanted in 2019 and the data gathered for this study was recorded across two different sessions. One in 2021 and the other in 2022.

| Bloc | MDV | p |
|---|---|---|
| CLIN-D1-B1 | 0.087 | 0.063 |
| CLIN-D1-B2 | 0.098 | 0.024 |
| CLIN-D1-B3 | 0.085 | 0.59 |
| CLIN-D1-B4 | 0.097 | 0.57 |
| CLIN-D1-B5 | 0.096 | 0.27 |
| CLIN-D1-B6 | 0.11 | 0.092 |
| CLIN-D2-B1 | 0.092 | 0.64 |
| CLIN-D2-B2 | 0.094 | 0.46 |
| CLIN-D2-B3 | 0.12 | 0.13 |
| P5-D1 | 0.092 | 0.64 |
| P5-D2 | 0.094 | 0.34 |
| P5-D3 | 0.12 | 0.73 |

**Table 1.** Acoustic contamination evaluation of the included blocks of recordings in this study. MDV stands for Mean Diagonal Value. For each block, the p-value corresponds to the probability that the observed MDV is higher than chance. Because these tests are repeated for the different blocks, these p-values need to be corrected for multiple comparisons in each participant. Using a Bonferroni correction, a 5% risk of wrongly rejecting the presence of acoustic contamination would thus correspond to $p < 0.05/9$ for CLIN and $p < 0.05/3$ for P5. At these thresholds, all blocks could thus be considered as not being contaminated.

### B. Task and data recording

Both participants were asked to read aloud sentences from the BY2014 corpus (https://zenodo.org/records/154083) [42] containing four sequences of vowels "/a/ /i/ /u/", "/y/ /e/ /ɛ/",

"/ə/ /o/ /ɑ̃/", "/ɔ̃/ /œ̃/"; as well as short French sentences. Each sequence of vowels or sentence was displayed multiple times on a screen in front of the participant which he/she read overtly. ECoG and audio signals were recorded synchronously.

The P5 dataset contained 634 sentences comprising 28 repetitions per sequence of vowels and 5 repetitions per complete sentence. On the other side CLIN dataset contained 1332 sentences comprising 63 repetitions per sequence of vowels and 8 repetitions per complete sentence. For P5 participant, ECoG and audio signals were recorded using a Blackrock Neuroport system at a sampling rate of 30 kHz (see ref [3] for details). For participant CLIN, neural signals were recorded on the 32 most ventral electrodes of the left implant at 585.6 Hz using the WIMAGINE system synchronized with a CED micro1401 system sampling the audio signals at 100 kHz.

### C. Data contamination analysis

The contamination of the neural signals by the audio recordings was evaluated on P5 and CLIN datasets using the pipeline described in Roussel et al 2020 [43]. The neural data was cropped to the speech moments where no artifacts were detected. Then the acoustic contamination was assessed by computing the cross-correlation matrix between the spectrograms of the audio and the neural signals over the frequency range 60-200 Hz. This led to the computation of the Mean Diagonal Value (MDV) of the matrix. Finally, each MDV was compared with 10000 surrogates to evaluate its statistical significance.

### D. Neural data preprocessing

Artifacts such as line noise were removed from neural signals using common average referencing. Then spectrograms were computed for each electrode signal using a Fast Fourier Transform with a moving hamming window of 200 ms and a 10-ms frame step. The power spectral density of each frequency band was averaged within 10 Hz bands between 0 and 200 Hz to end up with 20 spectral features sampled at 100 Hz for each electrode. Additionally, a Local Field Potential (LFP) feature was computed by filtering the raw signal at each electrode between 0.5 and 5 Hz. In total, there were thus 21 neural features per electrode (Figure 1.b.).

### D. Decoded target

As for the decoded target, we extracted from every recorded audio 25 Mel Cepstral features (Mel), 2 aperiodicity features (Ap) the fundamental frequency (F0). To better fit the F0 variations, the voicing (UV) was computed to be used for F0 clipping to voiced and unvoiced speech parts. The decoded acoustic features are fed to the WORLD vocoder [44] to reconstruct the relative speech spectrogram and the corresponding audio signal.

### E. General decoding model

The decoding process (Figure 1.b) is based on an encoder-decoder scheme addressing all sentences one by one globally (sentence-by-sentence decoding). At the encoding stage, two architectures were examined (Figure 2), a Convolutional Neural Network (CNN, Figure 2.a) and a Vision Transformer (ViT, Figure 2.b). The decoder is a bidirectional LSTM (bi-LSTM). This architecture receives as input the ECoG-derived neural features $e$ and generates a 29-dimensional time varying vector $\hat{a}$ fitting the target acoustic vector $a$ in a trial-by-trial basis through the minimization of the Mean Squared Error (MSE) Loss computed as follows:

$$MSELoss = \frac{1}{T*29}\sum_{t=1}^{T}\sum_{c=1}^{29}(a(t,c) - \hat{a}(t,c))^2, \quad (1)$$

where T is the number of samples of a sentence.

### F. Encoders architectures

For the CNN architecture, the input is structured as $[N_x, N_y, 21]$, where $N_x$ and $N_y$ represent the spatial dimensions of the electrode grid, 9×8 for P5 and 4×8 for CLIN. The CNN consists of three layers of 2D convolutions. Each layer uses convolutional kernels of size 3×3 and 2×2 for P5 and CLIN respectively. The number of kernels per layer is 128, 64, and 32. Each convolutional layer is followed by batch normalization and a Leaky ReLU activation with a negative slope of 0.2 (Figure 2.a).

In contrast, the ViT architecture flattens the input at each time step into vectors of size $N_f = [N_x \times N_y \times 21]$ thus 1512 and 672 for P5 and CLIN, respectively. This vector is fed to an MLP embedding it into a latent space through two projections: the first from the input dimension to 256 and the second from 256 to 176. Each of these projections is followed by a tanh activation. The resulting embeddings are temporally position-encoded with the sine-cosine method [45] summed with the embedded vectors and fed to a multi-head attention block with 4 heads each of dimension 32. The final step consists in a 2-layer MLP projector going from 176 to 64 (with a GeLU activation), then back to 176 (Figure 2.b). Unlike the CNN, which is participant-specific, the ViT architecture is shared across participants, differing only in the initial projection layer.

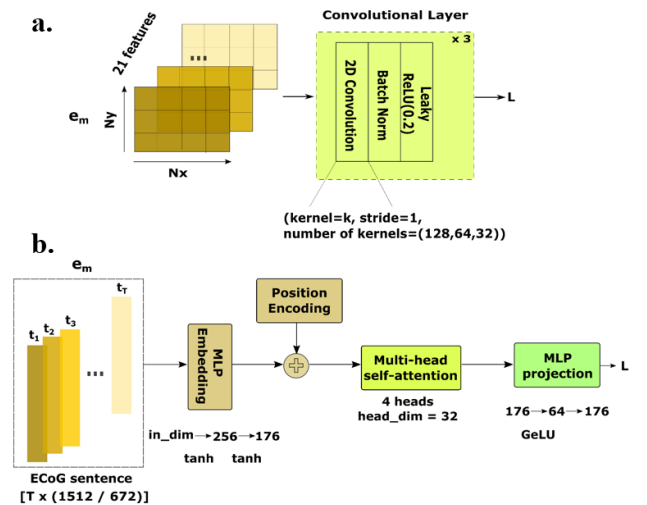

**Figure 2.** Encoders architectures. (a) presents the CNN architecture and (b) presents the ViT one.

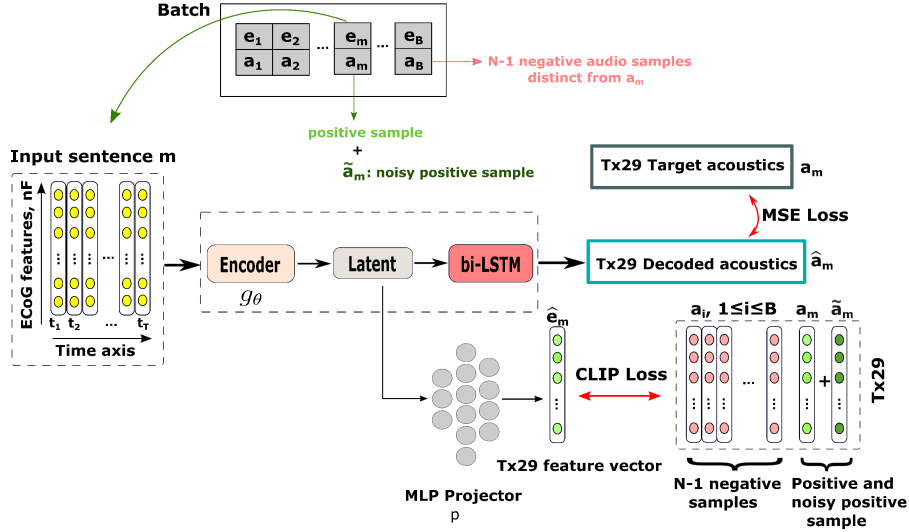

**Figure 3.** Incorporation of CLIP loss into the training process. The model is optimized through two parallel pathways: one combining the encoder and decoder with MSE loss, and another applied solely to the encoder with CLIP loss.

### G. Decoder architecture

The decoder consists of a three-layer bidirectional LSTM with a hidden size of 256. Its input is the latent time-varying vectors F output by the encoder, and its output is passed through an MLP to produce 29-dimensional target vectors. The MLP first projects the 512-dimensional LSTM output to 1024 units, followed by a 50% dropout layer, and then applies a final linear projection to obtain a 29-dimensional output vector $\hat{a}$ at each time step.

### H. Contrastive Learning

Both datasets utilized in this study are of a reduced size, challenging to the training process. To cope with this limitation, an additional learning phase is utilized leveraging contrastive training based on the CLIP loss [46] to train the encoder only (Figure 3). Thus, the training process comprises two losses, an MSE Loss and a CLIP Loss.
To formalize the contrastive process, two function blocks are introduced $g_\theta$ and p corresponding to the encoder and an MLP projector, respectively, and both are combined in the function $f_{\hat{\theta}}$ as follows:

$$f_{\hat{\theta}} = p \circ g_\theta \qquad (2)$$

At each training step, a batch of B sentences of ECoG $(e_m)_{1 \leq m \leq B}$ and corresponding audio representations $(a_m)_{1 \leq m \leq B}$ are utilized, where each acoustic vector $a_i$ corresponds to the ECoG $e_i$. Applying $f_{\hat{\theta}}$ on each input ECoG sentence $e_m$ generates $\hat{e}_m$, i.e. $f_{\hat{\theta}}(e_m) = \hat{e}_m$.
Training $f_{\hat{\theta}}$ results in training p and $g_\theta$, where only $g_\theta$ is retained for the inference as p is used only to map the encoders latent output into the same dimension as the acoustic feature vectors for the CLIP process during training.

To perform the contrastive training, N+1 samples are utilized comprising N-1 negative ones consisting in the acoustic vectors of the batch that correspond to sentences distinct from that of $e_m$. The batch is likely to comprise multiple occurrences of the same sentence or sequence of vowels. These repetitions are excluded from the set of negative samples. Then, two positive samples are used comprising the target acoustic vector $a_m$ corresponding to $e_m$ and a noisy version of $a_m$, noted $\tilde{a}_m$, computed in two steps. First, a *noise* vector sampled from the normal distribution is added to $a_m$:

$$a_{m\_noised} = a_m + noise \qquad (3)$$

Then, $\tilde{a}_m$ is computed through the following equation:

$$\tilde{a}_m = (a_{m\_noised} - mean(a_{m\_noised})) \\ * noise + mean(a_{m\_noised}) \qquad (4)$$

For the present notations, in the N+1 sampled candidates, $a_N = a_m$ and $a_{N+1} = \tilde{a}_m$ and the other indices from 1 to N-1 correspond to negative samples.
Subsequently distance matrix $V = (v_{mn})_{1 \leq m \leq B, 1 \leq n \leq N+1}$ is computed between each encoded and projected ECoG vector $\hat{e}_m$ and the N+1 sampled candidates using the MSE distance as follows:

$$v_{mn} = MSE(\hat{e}_m, a_n) \qquad (5)$$

Subsequently, the LogSoftmax function is applied over the matrix V, generating the matrix $\widetilde{V} = (\tilde{v}_{mn})_{1 \leq m \leq B, 1 \leq n \leq N+1}$ defined as follows:

$$\tilde{v}_{mn} = LogSoftmax(v_{mn}) \qquad (6) \\ = Log \frac{\exp(MSE(\hat{e}_m, a_n))}{\sum_{n=1}^{N+1} \exp(MSE(\hat{e}_m, a_n))}.$$

Finally, the CLIP loss to be optimized is given by:

$$CLIPLoss = -\sum_{m=1}^{B} \tilde{v}_{mN} + \tilde{v}_{mN+1}. \quad (7)$$

### I. Augmentation of the neural training corpus

As previously stated, the datasets of both patients comprise multiple repetitions per sentence and sequence of vowels, i.e. each sentence s of the corpus has k repetitions: $(e_s^i)_{1 \leq i \leq k}$. An augmentation consists of selecting a given ECoG trial $e_s^i$ mapping it to an audio $a_s^j$, $1 \leq i, j \leq k$, $i \neq j$. The temporal alignment between $a_s^i$ and $a_s^j$ is estimated using the dynamic time warping algorithm (DTW), used subsequently to align temporally $e_s^i$ on $a_s^j$ (Ben Ticha et al. [33]).
This operation allows the multiplication of the training set tremendously. Here, we restrain the multiplication size of the training sets to a factor of 4 on both P5 and CLIN corpuses.

### J. Model Training

To train the encoder-decoder architecture, the MSE Loss and the CLIP Loss are combined. The *MSE Loss* optimized both the encoder and the decoder, and the *CLIP Loss* optimized the encoder only. During the training phase, the sum of the two losses is computed and optimized:

$$Loss = MSE\ Loss + CLIP\ Loss. \quad (8)$$

The Adam optimizer was used without weight decay, with a learning rate of 0.0004 without a scheduler.

### K. Cross validation

The decoding scheme was based on 10-fold cross-validation. Each dataset was divided into (Train, Validation, Test) sets in the following proportions, (72%, 18%, 10%) corresponding to (457, 114, 63) and (960, 239, 133) for P5 and CLIN datasets, respectively. Thus, 10% of the dataset was predicted at each fold without any overlapping between test sets.

### L. Evaluation of metrics

To evaluate the learning process, three metrics are utilized. The Pearson Correlation Coefficient (PCC) between the 29 decoded and target acoustic vectors. The Mel-Cepstral Distortion (MCD) computed between the 25 target and decoded Mel coefficients, to assess objectively the intelligibility of the decoded audio. And the F1 score utilized to assess the correctness of vowel's decoding. It is computed in two steps. First each decoded vowel is compared with all the vowels in the corpus using the MCD distance. The vowel in the original corpus delivering the least MCD score is considered as the vowel decoded. Then the F1 score is deduced from this computation, using the following equation:

$$F1\_score = \frac{TP}{TP + \frac{1}{2} * (FP + FN)}, \quad (9)$$

where TP, FP and FN correspond to True Positives, False Positives and False Negatives, recursively.

### E. Early Stopping

To limit overfitting, early stopping is utilized during the training phase using the MCD scores computed on the validation set with a patience of 20. If the MCD does not improve, the stopping counter is incremented by one and the model parameters are not saved. Alternatively, the stopping counter is set to 0 and the model parameters are saved.

### F. Saliency maps

Saliency maps are computed as part of an explainability task evaluating the contributions of the different ECoG inputs $e = (e_{i,t})_{1 \leq i \leq Nf,\ 1 \leq t \leq T}$ to the MSE Loss computed between target and decoded acoustic vectors. The adopted method is based on the differentiation of the MSE Loss relative to the ECoG as follows:

$$S = \frac{\partial MSE\ Loss}{\partial e_{it}}. \quad (10)$$

However, this computation is likely to generate highly noisy maps [4]. To reduce this noise and to obtain more robust computations, the SmoothGrad method [47] is utilized. First noise vectors are sampled n times from a gaussian distribution $N(0, \hat{\sigma}^2)$, where $\hat{\sigma} = \frac{\sigma}{e_{max} - e_{min}}$, $\sigma = 0.15$, $e_{max}$ and $e_{min}$ are the maximum and minimum values of $e$ respectively ($e$ is non-null and non-constant). Our saliency maps $\hat{S}$ are computed as follows:

$$\hat{S} = \frac{1}{n} \sum_{i=1}^{n} S(e + noise_i), n = 50. \quad (11)$$

## III. RESULTS

All subsequent results were obtained with neural corpuses evaluated where a contamination analysis was computed. No contamination was observed on the neural corpuses (Table 1).

### A. The ViT encoder outperforms the CNN encoder

The ViT setup consistently outperformed the CNN one across all evaluation metrics and both datasets (Figure 4). For P5, the ViT achieved a mean PCC of 0.479 ± 0.114, compared to 0.426 ± 0.109 for the CNN. The ViT also demonstrated lower MCD (0.4119 ± 0.726 vs. 0.4362 ± 0.682) and higher F1 score (0.16 vs. 0.12). This result was also observed on the CLIN dataset:

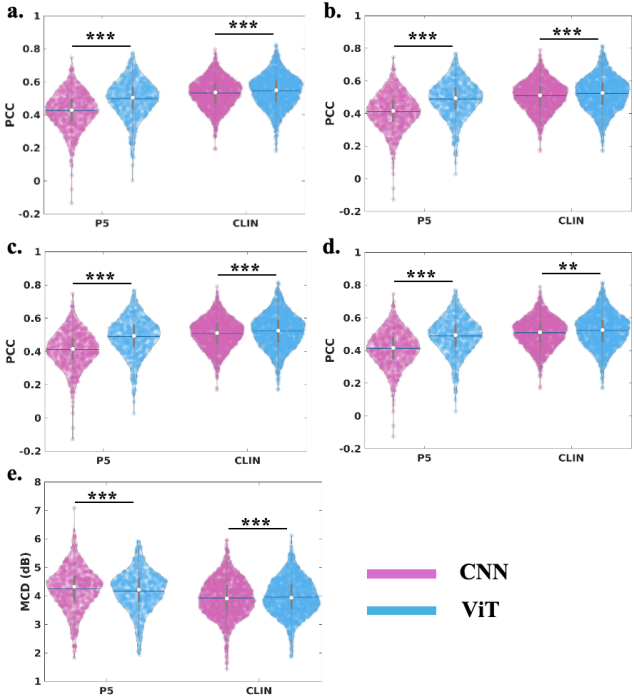

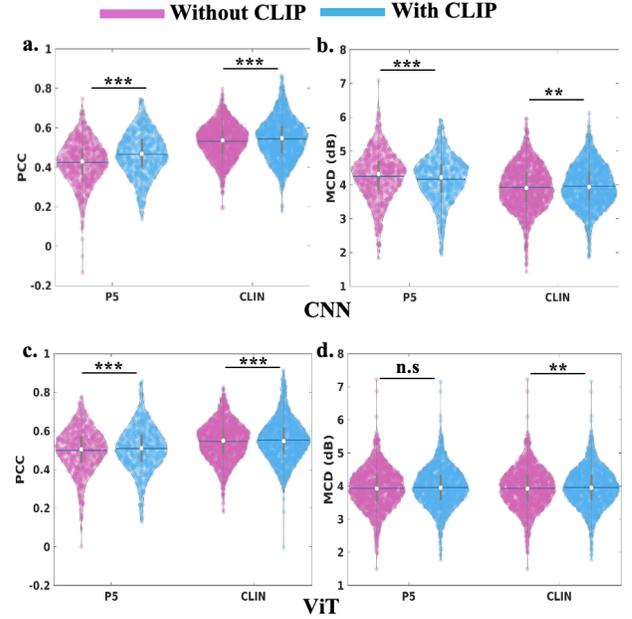

**Figure 4. Comparison of decoding performance between CNN and ViT-based models on the P5 and CLIN corpora.** (a) Pearson correlation coefficients (PCC) between decoded and target acoustic features. (b–d) PCC for Mels, aperiodicities and F0, respectively. (d) Mel-cepstral distortion (MCD) between decoded and target Mels. Statistical significance was evaluated using the Wilcoxon signed-rank test with Bonferroni correction. ***: $p < 0.001$, **: $p < 0.01$.

the ViT setup achieved a PCC of $0.546 \pm 0.095$ compared to $0.532 \pm 0.084$ for the CNN one, with lower MCD ($3.869 \pm 0.571$ vs. $3.973 \pm 0.571$). As illustrated in Figure 4, all these differences were significant (either $p < 0.01$ or $p < 0.001$; Wilcoxon rank sum test with Bonferroni correction). Regarding the F1 score, it also improved using ViT (0.20 vs. 0.15).

### B. Adding CLIP training improves decoding

As shown in Figure 5, the CLIP joint training approach (*MSE Loss + CLIP loss*) improved the mean PCC for both models. The ViT+CLIP setup reached a PCC of $0.518 \pm 0.1$ on P5 and $0.557 \pm 0.089$ (mean ± SD) on CLIN, and the MCD decreased on P5 (to $3.929 \pm 0.636$) but showed a slight increase on CLIN. The F1 score improved with the CNN setup, slightly dropped for ViT on P5, and remained stable for CLIN. Overall, ViT+CLIP outperformed CNN+CLIP.

### C. Data-Driven Neural Variability-Based augmentation improves decoding

Utilizing our proposed neural data augmentation strategy leveraging the intrinsic variability in repeated neural recordings, the metrics improved globally (Figure 6). In particular, the best-performing setup (ViT+CLIP) was further enhanced with this augmentation approach. On P5, ViT+CLIP+Augmentation achieved a PCC of $0.529 \pm 0.105$ and an MCD of $3.926 \pm 0.686$. On CLIN, the mean PCC rose to $0.563 \pm 0.1$ and the MCD decreased to $3.836 \pm 0.686$. The F1 scores improved to 0.29 (P5) and 0.26 (CLIN). These results affirm the effectiveness of the proposed augmentation method and the consistent superiority of the ViT architecture over the CNN one.

**Figure 5.** Decoded performances evolution when CLIP is used or not for the CNN setups (a and b), and the ViT one (c and d). Statistical tests were performed using Wilcoxon signed rank test with Bonferroni correction. ***: $p < 0.001$, **: $p < 0.01$, n.s.: $p > 0.05$.

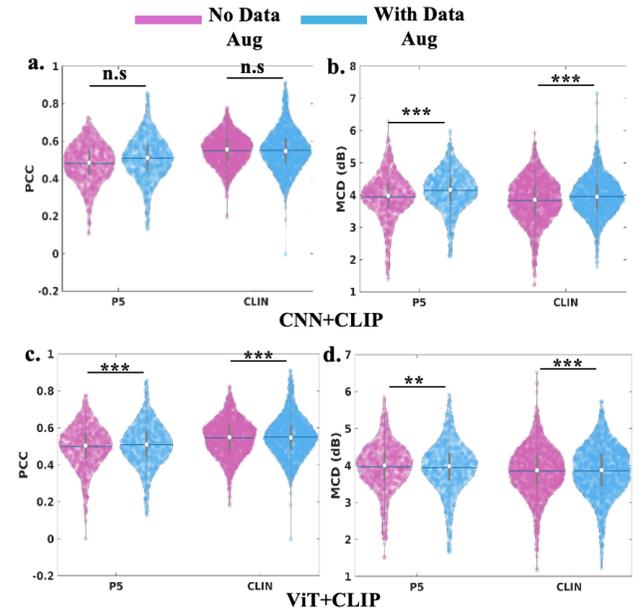

**Figure 6.** Evolution of PCC and MCD performance with or without data augmentation in CLIP training for CNN (a, b) and ViT (c, d) models, and the impact of transfer learning in the ViT+CLIP with data augmentation setup (e, f). Statistical significance was assessed using the Wilcoxon signed-rank test for (a–d) and the Quade–Conover paired test for (e, f), with Bonferroni correction. ***: $p < 0.001$, **: $p < 0.01$, n.s.: $p > 0.05$.

### D. Transfer Learning Across Participants improves decoding

Due to patient-specific configurations in the CNN architecture (Figure 2.a), transfer learning with this setup was not considered. However, in the ViT setup, only the first projection layer is patient-specific, and the remainder is shared. We thereby applied transfer learning using the ViT+CLIP+Augmentation model. Transfer learning improved all metrics, especially for P5: PCC rose to $0.564 \pm 0.109$ (from $0.529 \pm 0.105$), MCD decreased to $3.777 \pm 0.736$ (from $3.926 \pm 0.686$) (Figure 7), and the F1 score increased from 0.29 to 0.43. Improvements on CLIN were more modest but consistent. Table 2 shows a summary of the performances of the different setups on P5 and CLIN datasets. Figure 9 shows an example of a successful reconstruction of a sequence of vowels for each participant (see also the corresponding Supplementary audio files 1-8).

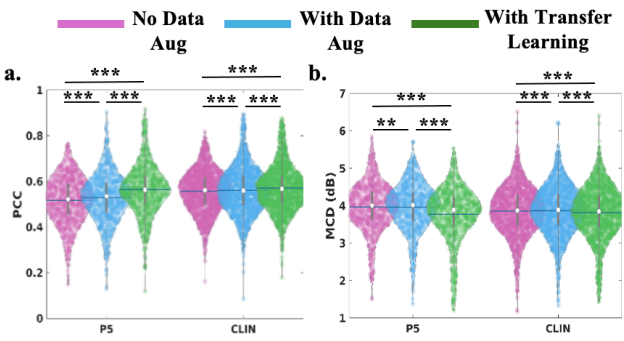

**Figure 7.** Evolution of PCC and MCD performance with or without data augmentation in CLIP training for CNN (a, b) and ViT (c, d) models, and the impact of transfer learning in the ViT+CLIP with data augmentation setup (e, f). Statistical significance was assessed using the Wilcoxon signed-rank test for (a–d) and the Quade–Conover paired test for (e, f), with Bonferroni correction. ***: p < 0.001, **: p < 0.01.

### E. Vowel decoding accuracy

Figure 8 illustrates the confusion matrices of the decoded vowels regarding these best performing setups, showing that decoding was better for P5 (Figure 8a) than for CLIN (Figure 8c). To assess that the decoding performance was greater than chance, we repeated the training 10 times with shuffled targets, and found that in this case the decoding performance was indeed lower than that obtained with unshuffled labels (p<0.005, Mattney-Whitney U test) for both P5 (Figure 8b) and CLIN (Figure 8d).

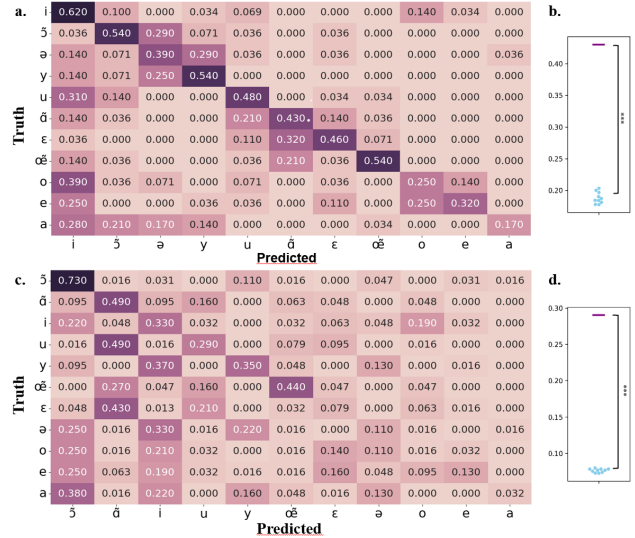

**Figure 8.** Confusion matrices of the decoded vowels from P5 (a) and CLIN (c) using ViT+CLIP+Data augmentation (Aug) and transfer learning (TL). b and d show the F1-scores computed using the true data labels (purple) and the shuffled labels across 10 repetitons (blue), for P5 and CLIN, respectively. The statistical test was evaluated with Mann–Whitney U test, ***: p<0.005.

### F. Evaluation of Electrode and Neural Feature Contributions

To understand the contributions of the spatial and spectral input features to the decoding performance, saliency maps were computed using the best-performing setup: ViT+CLIP+Data Augmentation+Transfer Learning.

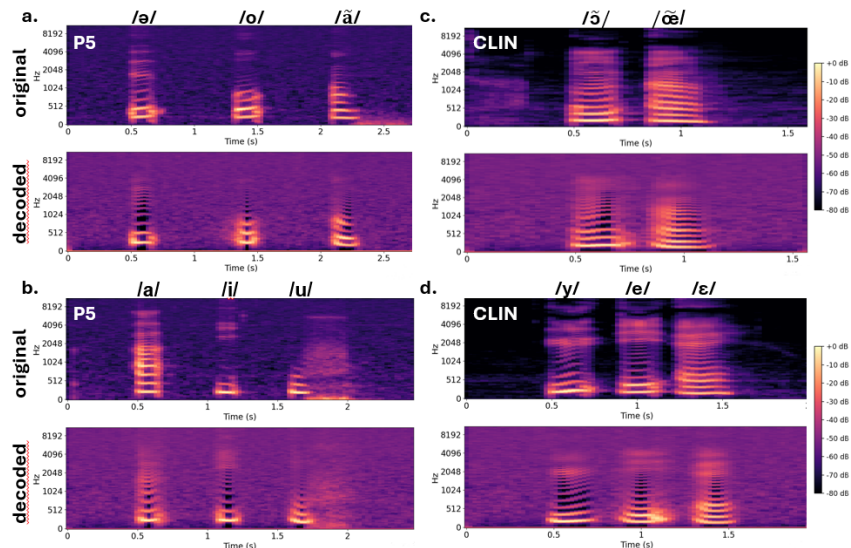

**Figure 9.** Examples of successfully reconstructed spectrograms of four sequences of vowels. Two for participant P5 (left column) and two for participant CLIN (right column). Top rows: Original speech sound. Rows 2: sound reconstructed with the ViT + CLIP + Data augmentation + Transfer learning setup.

|  | Avg PCC | | Avg MCD | | F1_score | |
| --- | --- | --- | --- | --- | --- | --- |
|  | P5 | CLIN | P5 | CLIN | P5 | CLIN |
| CNN | 0.426±0.109 | 0.532±0.084 | 4.362±0.682 | 3.973±0.571 | 0.12 | 0.15 |
| CNN+CLIP | 0.482±0.105 | 0.551±0.077 | 4.194±0.639 | 3.996±0.566 | 0.13 | 0.18 |
| CNN+CLIP+Aug | 0.489±0.105 | 0.549±0.095 | 4.166±0.665 | 3.926±0.686 | 0.19 | 0.21 |
| ViT | 0.479±0.114 | 0.546±0.095 | 4.119±0.526 | 3.869±0.642 | 0.16 | 0.20 |
| ViT+CLIP | 0.518±0.100 | 0.557±0.089 | 3.929±0.650 | 3.886±0.636 | 0.14 | 0.20 |
| ViT+CLIP+Aug | 0.529±0.105 | 0.563±0.100 | 3.926±0.686 | 3.836±0.686 | 0.29 | 0.26 |
| ViT+CLIP+Aug+TL | 0.564±0.109 | 0.570±0.100 | 3.777±0.736 | 3.814±0.689 | 0.43 | 0.29 |

**Table 2.** Summary table of the difference setups performances assessed on P5 and CLIN datasets. Aug refers to data augmentation and TL to transfer learning.

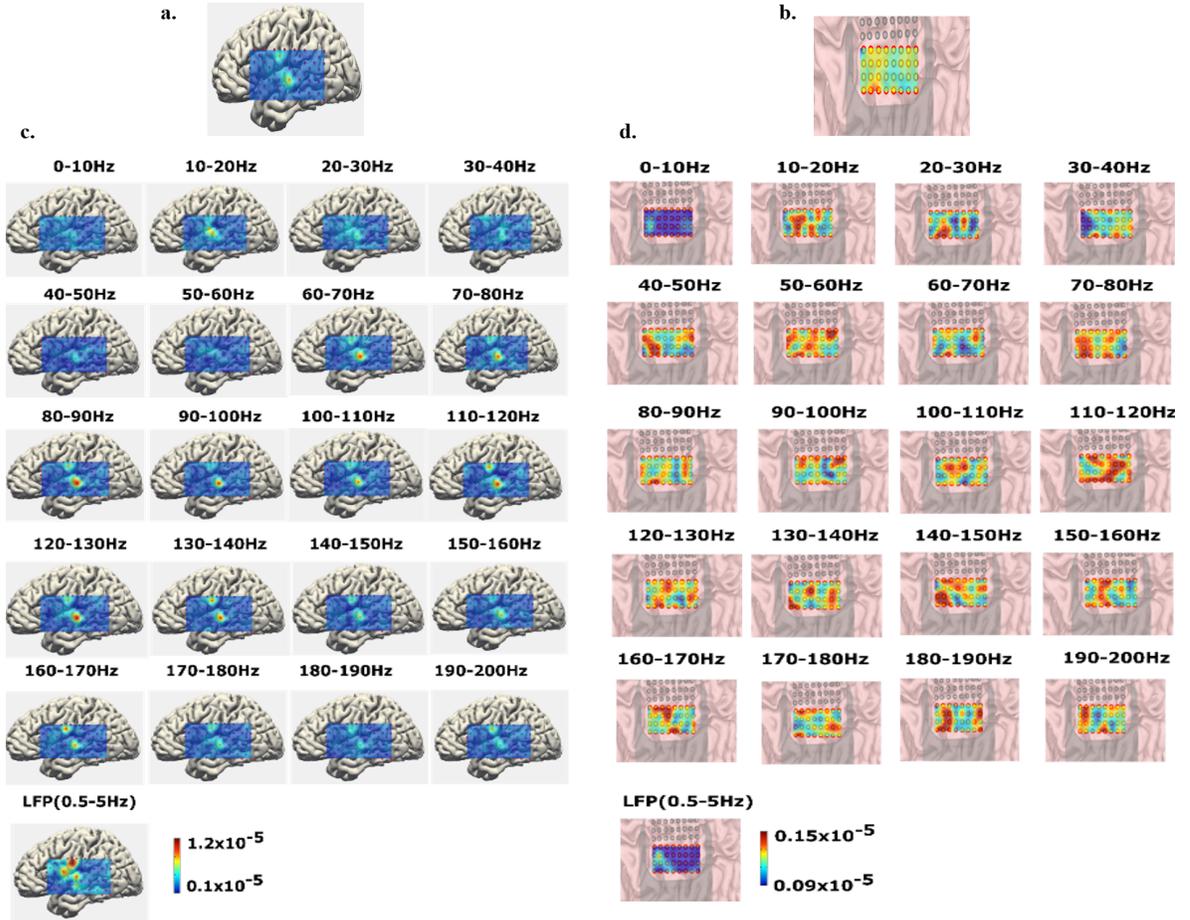

**Figure 10.** Electrodes and features contributions maps of P5 (Figure a and c respectively) and CLIN (Figure b and d respectively).

**Electrode Contributions**

Examining the electrodes contributions is informative of the cortical areas involved in speech decoding. For participant P5, the saliency map indicated strong contributions from the Superior Temporal Gyrus (STG), and the ventral somatosensory cortex (vSMC) (Figure 10a), aligning with the literature regarding speech cortical areas. In contrast, for participant CLIN, the saliency maps highlighted contributions predominantly from the motor cortex, particularly the most ventral motor electrodes (Figure 10b).

**Neural Features contributions**

Spectral saliency maps for P5 reveal contributions of the low beta band (10-20 Hz), gamma and high gamma bands (60 – 200 Hz) and of the slow LFP (Figure 10c). Conversely, CLIN saliency map showed a uniform contribution of 10 to 200 Hz feature bands with a minimal contribution of 0 to 10 Hz and the slow LFP features (Figure 10d).

## IV. Discussion

In this study we demonstrated an application of an encoder-decoder pipeline for the direct decoding of overt speech from ECoG neural data. The encoder is provided with neural features extracted from the neural signals and the decoder generates acoustic vectors which are subsequently fed to a vocoder. Two types of encoders were evaluated, a CNN and a ViT. The results demonstrate that the ViT encoder outperforms the CNN one across all metrics on two participants datasets P5 and CLIN. This finding indicates the potential benefit of incorporating transformer-based architectures into speech decoding pipelines [24], [33]. Indeed, there is a need to enhance the performance of the utilized speech decoding models, particularly in the context of continuous decoding frameworks, with the objective of improving the intelligibility of the decoded speech. Previously, a ViT was employed for speech decoding in an encoder-decoder configuration, analogous to our own approach, for a classification task [48] or a regression task [49], [50]. These studies demonstrated that the ViT setups consistently outperformed bi-LSTM architectures. In the present decoding scheme, a contrastive learning-based approach was also incorporated. Such approach has been proved as a powerful means to pretrain models for downstream tasks. It is based on the idea of minimizing distances between positive samples and maximizing distances between positive and negative ones. Particularly the CLIP model was designed to compare between two modality distributions of input data. In [35] a CLIP model has been utilized to classify perceived speech items from EEG and MEG brain activity. Based on this process, high accuracy was obtained on MEG dataset with 8 different classes. In our case where the task is generative, we incorporated the CLIP training into a continuous regression process. This increased the performance of the decoding algorithm. Then, to further improve decoding accuracy, a neural variability-based data augmentation approach was implemented. It allowed to increase the performance of both encoders' setups on P5 and CLIN datasets with a more notable increase on P5. This might suggest that an increase of P5 dataset size would increase the model's performance. Finally, we took advantage of the fact that the ViT architecture was largely shared between participants except for the first projection layer. This allowed performing transfer learning across subjects, which was evaluated on the ViT+CLIP setup with data augmentation.

Interestingly, this operation improved decoding performances on both P5 and CLIN datasets, although one deals with epidural data and the other subdural data that cover distinct brain areas. This suggests that transfer learning between two distinct neural corpuses can still increase the decoding performances even when the datasets are not of the same size and nature.

With the best performing decoding pipeline, ViT+CLIP with data augmentation and transfer learning, F1 score in decoding individual vowels from the vowel sequences reconstructions were well above chance level for both participants (Figure 8). This results suggests that beyond classification-based speech BCI approaches that have been used over the past years [17], [19], [22], [20], [23], [21], [51], [52], direct regression approaches that reconstruct the acoustic speech content without language models, as recently introduced with intracortical recordings [24], might also successfully extend to ECoG signals.

In the present study, the classification of vowels was performed using the best MCD score computed over the intervals delimitating each vowel reconstructed by regression. We found that, while the global PCC and MCD computed over the whole sentences were better for CLIN than P5, the F1 score was higher for P5. It is possible that this is due to the fact that the PCC and MCD scores were computed over the whole sentences or vowel sequences, including silences. In this case these scores are largely governed by the succession of speech and non-speech intervals, and less by the precise content of the speech-only intervals. Because the CLIN dataset contained more trials, the models might thus be more inclined to correctly reconstruct the speech/non speech sequences and thus to achieve higher global PCC and MCD scores. Yet, when focusing only on the speech intervals to classify the vowels, the MCD was better predicted for P5 who had more channels which were better located over speech-engaged regions.

Saliency maps were computed for the best decoding pipeline to identify the brain areas as well as the neural features most relevant to speech decoding. Regarding P5, the most informative electrodes were located over the STG and the vSMC. The contribution of the vSMC showed that the activation tended to extend more dorsally. These findings align with those reported in [15], where STG and vSMC were also the primary brain regions contributing to the decoding of non-speech acoustics using a non-causal decoding scheme. However, for causal decoding, the STG was found to be non-participatory, with only the vSMC involved in the decoding process. This observation merits consideration in future training processes, as a speech BCI would ideally use a causal decoder suitable for real-time applications.

In contrast, for CLIN participant, the electrodes identified as the most contributory to the decoding process were those located most ventral over the region of the hand motor cortex covered by the implant, thus closest to the dLMC area, a key region controlling phonation [41].

Afterwards, the contributions of each of the 21 utilized neural features were examined for both participants. For P5, the most active spots in the contribution maps are observed for features ranging from 60-70 Hz (included in the gamma band range) to 190-200Hz, the 70-200 Hz features represent the high-gamma band activity. Activity related to these features is highlighted in the STG, and the vSMC. This observation is expected and is consistent with previous studies consistently using high-gamma signals for speech decoding. Interestingly the slow LFP feature also demonstrated a predominant contribution compared to the other neural features for the decoding of all acoustic coefficients. This was the case in the STG, the MTG, the vSMC and to a lesser extent to Broca's area, and aligns with previous studies using not only high gamma neural features ([9], [12], [17], [18], [19], [53]), but also an LFP feature to improve decoding (0.3-17 Hz in [19], 0-30 Hz in [11] and 0.3-100 Hz in [18], [53]). For CLIN participant, all neural frequency features between 10 and 200 Hz contributed to the decoding process, while the LFP feature did not exhibit notable contribution, which could be related to the non-optimal location of the ECoG grid in CLIN.

In conclusion, the present study proposes a ViT-based encoder-decoder pipeline incorporating contrastive learning for continuous direct speech decoding from ECoG data. Furthermore, it is to our knowledge the first attempt to decode speech from epidural neural recordings. Above chance decoding could be obtained despite a limited number of electrodes not optimally positioned over the main motor speech areas. These results thus encourage future speech BCI studies using the epidural WIMAGINE technology.

ACKNOWLEDGMENTS

We thank P5 and CLIN for their time and active participation to the experiments. We also would like to thank the multidisciplinary technical and clinical teams at Clinatec (CEA-LETI and CHU-Grenoble Alpes) for their involvement in the BCI&Tetraplegia clinical trial (NCT02550522). Finally, we thank Vincent Auboiroux for his assistance in providing medical reconstruction images of the implant location. This study was supported in part by the FRM foundation under grant No DBS20140930785 and the PhD grant of GS, the European Union's Horizon 2020 research and innovation program under Grant Agreements No. 732032 (BrainCom), the French National Research Agency under Grant Agreement No. ANR-16-CE19-0005-01 (Brainspeak) and ANR-20-CE45-0005 (BrainNet) and "Investissements d'avenir" program (ANR-15-IDEX-02) funding the Grenoble-Neurotech project. It was also supported by CEA recurrent funding, the Carnot Institute CEA-Leti and the French Ministry of Health and Research (Grant PHRC 15-124), and by the Zhejiang Provincial Key Research and Development Program (2021C03003) and through the China Scholarship Council (201906320337).

REFERENCES

[1] J. S. Brumberg, A. Nieto-Castanon, P. R. Kennedy, and F. H. Guenther, "Brain-Computer Interfaces for Speech Communication," *Speech Commun*, vol. 52, no. 4, pp. 367–379, 2010, doi: 10.1016/j.specom.2010.01.001.

[2] F. Bocquelet, T. Hueber, L. Girin, S. Chabardes, and B. Yvert, "Key considerations in designing a speech brain-computer interface," *J. Physiol. Paris*, vol. 110, no. 4, pp. 392–401, 2016, doi: 10.1016/j.jphysparis.2017.07.002.

[3] A. B. Silva, K. T. Littlejohn, J. R. Liu, D. A. Moses, and E. F. Chang, "The speech neuroprosthesis," *Nat. Rev. Neurosci.*, vol. 25, no. 7, pp. 473–492, July 2024, doi: 10.1038/s41583-024-00819-9.

[4] K. E. Bouchard, N. Mesgarani, K. Johnson, and E. F. Chang, "Functional organization of human sensorimotor cortex for speech articulation," *Nature*, vol. 495, no. 7441, pp. 327–332, 2013, doi: nature11911%20%5Bpii%5D%2010.1038/nature11911.

[5] C. Cheung, L. S. Hamiton, K. Johnson, and E. F. Chang, "The auditory representation of speech sounds in human motor cortex," *eLife*, vol. 5, pp. 1–19, 2016, doi: 10.7554/eLife.12577.

[6] S. Martin *et al.*, "Decoding spectrotemporal features of overt and covert speech from the human cortex.," *Front. Neuroengineering*, vol. 7, no. May, p. 14, Jan. 2014, doi: 10.3389/fneng.2014.00014.

[7] E. M. Mugler *et al.*, "Direct classification of all American English phonemes using signals from functional speech motor cortex," *J. Neural Eng.*, vol. 11, no. 3, p. 035015, June 2014, doi: 10.1088/1741-2560/11/3/035015.

[8] N. F. Ramsey, E. Salari, E. J. Aarnoutse, M. J. Vansteensel, M. G. Bleichner, and Z. V. Freudenburg, "Decoding spoken phonemes from sensorimotor cortex with high-density ECoG grids," *NeuroImage*, vol. 180, no. April 2017, pp. 301–311, 2018, doi: 10.1016/j.neuroimage.2017.10.011.

[9] M. Angrick *et al.*, "Speech synthesis from ECoG using densely connected 3D convolutional neural networks," *J. Neural Eng.*, 2019, doi: 10.1088/1741-2552/ab0c59.

[10] C. Herff *et al.*, "Generating Natural, Intelligible Speech From Brain Activity in Motor, Premotor, and Inferior Frontal Cortices," *Front. Neurosci.*, vol. 13, no. November, pp. 1–11, 2019, doi: 10.3389/fnins.2019.01267.

[11] G. K. Anumanchipalli, J. Chartier, and E. F. Chang, "Speech synthesis from neural decoding of spoken sentences," *Nature*, vol. 568, pp. 493–98, 2019, doi: 10.1101/481267.

[12] J. G. Makin, D. A. Moses, and E. F. Chang, "Machine translation of cortical activity to text with an encoder–decoder framework," *Nat. Neurosci.*, 2020, doi: 10.1038/s41593-020-0608-8.

[13] S. Duraivel *et al.*, "High-resolution neural recordings improve the accuracy of speech decoding," *Nat. Commun.*, vol. 14, no. 1, p. 6938, Nov. 2023, doi: 10.1038/s41467-023-42555-1.

[14] J. Berezutskaya, Z. V. Freudenburg, M. J. Vansteensel, E. J. Aarnoutse, N. F. Ramsey, and M. A. J. van Gerven, "Direct speech reconstruction from sensorimotor brain activity with optimized deep learning models," *J. Neural Eng.*, 2023, [Online]. Available: http://iopscience.iop.org/article/10.1088/1741-2552/ace8be

[15] X. Chen *et al.*, "A neural speech decoding framework leveraging deep learning and speech synthesis," *Nat. Mach. Intell.*, vol. 6, no. 4, pp. 467–480, Apr. 2024, doi: 10.1038/s42256-024-00824-8.

[16] D. A. Moses, M. K. Leonard, J. G. Makin, and E. F. Chang, "Real-time decoding of question-and-answer speech dialogue using human cortical activity," *Nat. Commun.*, vol. 10, p. 3096, 2019, doi: 10.1038/s41467-019-10994-4.

[17] D. A. Moses *et al.*, "Neuroprosthesis for Decoding Speech in a Paralyzed Person with Anarthria," *N. Engl. J. Med.*, vol. 385, no. 3, pp. 217–227, 2021, doi: 10.1056/nejmoa2027540.

[18] S. L. Metzger *et al.*, "Generalizable spelling using a speech neuroprosthesis in an individual with severe

limb and vocal paralysis," *Nat. Commun.*, vol. 13, no. 1, pp. 1–15, 2022, doi: 10.1038/s41467-022-33611-3.

[19] S. L. Metzger *et al.*, "A high-performance neuroprosthesis for speech decoding and avatar control," *Nature*, vol. 620, no. 7976, pp. 1037–1046, 2023, doi: 10.1038/s41586-023-06443-4.

[20] S. Luo *et al.*, "Stable Decoding from a Speech BCI Enables Control for an Individual with ALS without Recalibration for 3 Months," *Adv. Sci.*, vol. 10, no. 35, p. 2304853, 2023, doi: 10.1002/advs.202304853.

[21] K. T. Littlejohn *et al.*, "A streaming brain-to-voice neuroprosthesis to restore naturalistic communication," *Nat. Neurosci.*, vol. 28, no. 4, pp. 902–912, Apr. 2025, doi: 10.1038/s41593-025-01905-6.

[22] F. Willett *et al.*, "A high-performance speech neuroprosthesis.," *Nature*, vol. 620, pp. 1031–36, 2023, doi: 10.1038/s41586-023-06377-x.

[23] N. S. Card *et al.*, "An Accurate and Rapidly Calibrating Speech Neuroprosthesis," *N. Engl. J. Med.*, vol. 391, no. 7, pp. 609–618, Aug. 2024, doi: 10.1056/NEJMoa2314132.

[24] M. Wairagkar *et al.*, "An instantaneous voice-synthesis neuroprosthesis," *Nature*, vol. 644, no. 8075, pp. 145–152, Aug. 2025, doi: 10.1038/s41586-025-09127-3.

[25] J. C. Barrese *et al.*, "Failure mode analysis of silicon-based intracortical microelectrode arrays in non-human primates.," *J. Neural Eng.*, vol. 10, no. 6, p. 066014, Dec. 2013, doi: 10.1088/1741-2560/10/6/066014.

[26] J. C. Barrese, J. Aceros, and J. P. Donoghue, "Scanning electron microscopy of chronically implanted intracortical microelectrode arrays in non-human primates," *J. Neural Eng.*, vol. 13, no. 2, p. 026003, 2016, doi: 10.1088/1741-2560/13/2/026003.

[27] X. Chen *et al.*, "Chronic stability of a neuroprosthesis comprising multiple adjacent Utah arrays in monkeys," *J. Neural Eng.*, vol. 20, no. 3, p. 036039, 2023, doi: 10.1088/1741-2552/ace07e.

[28] N. V. Hahn, E. Stein, and F. R. Willett, "Long-term performance of intracortical microelectrode arrays in 14 BrainGate clinical trial participants".

[29] M. Angrick *et al.*, "Real-time synthesis of imagined speech processes from minimally invasive recordings of neural activity," *Commun. Biol.*, vol. 4, no. 1, p. 1055, Sept. 2021, doi: 10.1038/s42003-021-02578-0.

[30] J. Kohler *et al.*, "Synthesizing Speech from Intracranial Depth Electrodes using an Encoder-Decoder Framework," *Neurons Behav. Data Anal. Theory*, vol. 6, no. 1, Dec. 2022, doi: 10.51628/001c.57524.

[31] K. A. Sillay *et al.*, "Long-Term Measurement of Impedance in Chronically Implanted Depth and Subdural Electrodes During Responsive Neurostimulation in Humans," *Brain Stimulat.*, vol. 6, no. 5, pp. 718–726, Sept. 2013, doi: 10.1016/j.brs.2013.02.001.

[32] M. J. Vansteensel *et al.*, "Longevity of a Brain–Computer Interface for Amyotrophic Lateral Sclerosis," *N. Engl. J. Med.*, vol. 391, no. 7, pp. 619–626, Aug. 2024, doi: 10.1056/NEJMoa2314598.

[33] M. B. Ben Ticha *et al.*, "A Vision Transformer Architecture For Overt Speech Decoding From ECoG Data," in *2024 46th Annual International Conference of the IEEE Engineering in Medicine and Biology Society (EMBC)*, July 2024, pp. 1–4. doi: 10.1109/EMBC53108.2024.10781877.

[34] J. Chen *et al.*, "Subject-Agnostic Transformer-Based Neural Speech Decoding from Surface and Depth Electrode Signals," Mar. 14, 2024, *Neuroscience*. doi: 10.1101/2024.03.11.584533.

[35] A. Défossez, C. Caucheteux, J. Rapin, O. Kabeli, and J.-R. King, "Decoding speech perception from non-invasive brain recordings," *Nat. Mach. Intell.*, vol. 5, no. 10, pp. 1097–1107, Oct. 2023, doi: 10.1038/s42256-023-00714-5.

[36] C. Mestais, G. Charvet, F. Sauter-Starace, M. Foerster, D. Ratel, and A. L. Benabid, "WIMAGINE®: Wireless 64-channel ECoG recording implant for long term clinical applications," *IEEE Trans. Neural Syst. Rehabil. Eng.*, vol. 23, no. 1, pp. 10–21, 2015, doi: 10.1109/TNSRE.2014.2333541.

[37] A. L. Benabid *et al.*, "An exoskeleton controlled by an epidural wireless brain–machine interface in a tetraplegic patient: a proof-of-concept demonstration," *Lancet Neurol.*, vol. 18, no. 12, pp. 1112–1122, 2019, doi: 10.1016/S1474-4422(19)30321-7.

[38] H. Lorach *et al.*, "Walking naturally after spinal cord injury using a brain–spine interface," *Nature*, vol. 618, no. 7963, pp. 126–133, 2023, doi: 10.1038/s41586-023-06094-5.

[39] F. Sauter-Starace *et al.*, "Long-Term Sheep Implantation of WIMAGINE®, a Wireless 64-Channel Electrocorticogram Recorder," *Front. Neurosci.*, vol. 13, p. 847, 2019, doi: 10.3389/fnins.2019.00847.

[40] C. Larzabal *et al.*, "Long-term stability of the chronic epidural wireless recorder WIMAGINE in tetraplegic patients," *J. Neural Eng.*, vol. 18, no. 5, p. 56026, 2021, doi: 10.1088/1741-2552/ac2003.

[41] B. K. Dichter, J. D. Breshears, M. K. Leonard, and E. F. Chang, "The Control of Vocal Pitch in Human Laryngeal Motor Cortex," *Cell*, vol. 174, no. 1, pp. 21-31.e9, June 2018, doi: 10.1016/j.cell.2018.05.016.

[42] F. Bocquelet, T. Hueber, L. Girin, C. Savariaux, and B. Yvert, "Real-Time Control of an Articulatory-Based Speech Synthesizer for Brain Computer Interfaces," *PLOS Comput. Biol.*, vol. 12, no. 11, p. e1005119, Nov. 2016.

[43] P. Roussel *et al.*, "Observation and assessment of acoustic contamination of electrophysiological brain signals during speech production and sound perception," *J. Neural Eng.*, vol. 17, no. 5, p. 056028, Oct. 2020, doi: 10.1088/1741-2552/abb25e.

[44] M. Morise, F. Yokomori, and K. Ozawa, "WORLD: A Vocoder-Based High-Quality Speech Synthesis System for Real-Time Applications," *IEICE Trans. Inf. Syst.*, vol. E99.D, no. 7, pp. 1877–1884, 2016, doi: 10.1587/transinf.2015EDP7457.

[45] A. Vaswani *et al.*, "Attention is All you Need".

[46] A. Radford *et al.*, "Learning Transferable Visual Models From Natural Language Supervision".

[47] D. Smilkov, N. Thorat, B. Kim, F. Viégas, and M. Wattenberg, "SmoothGrad: removing noise by adding noise," June 12, 2017, *arXiv*: arXiv:1706.03825. doi: 10.48550/arXiv.1706.03825.

[48] S. Komeiji *et al.*, "Transformer-Based Estimation of Spoken Sentences Using Electrocorticography," in *ICASSP 2022 - 2022 IEEE International Conference on Acoustics, Speech and Signal Processing (ICASSP)*, Singapore, Singapore: IEEE, May 2022, pp. 1311–1315. doi: 10.1109/ICASSP43922.2022.9747443.

[49] K. Shigemi *et al.*, "Synthesizing Speech from ECoG with a Combination of Transformer-Based Encoder and Neural Vocoder," in *ICASSP 2023 - 2023 IEEE International Conference on Acoustics, Speech and Signal Processing (ICASSP)*, Rhodes Island, Greece: IEEE, June 2023, pp. 1–5. doi: 10.1109/ICASSP49357.2023.10097004.

[50] S. Komeiji *et al.*, "Feasibility of decoding covert speech in ECoG with a Transformer trained on overt speech," *Sci. Rep.*, vol. 14, no. 1, p. 11491, May 2024, doi: 10.1038/s41598-024-62230-9.

[51] J. J. Jude *et al.*, "Decoding intended speech with an intracortical brain-computer interface in a person with longstanding anarthria and locked-in syndrome," *bioRxiv*, 2025, doi: 10.1101/2025.08.12.668516.

[52] E. M. Kunz *et al.*, "Inner speech in motor cortex and implications for speech neuroprostheses," *Cell*, vol. 188, no. 17, pp. 4658-4673.e17, Aug. 2025, doi: 10.1016/j.cell.2025.06.015.

[53] A. B. Silva *et al.*, "A bilingual speech neuroprosthesis driven by cortical articulatory representations shared between languages," *Nat. Biomed. Eng.*, vol. 8, no. 8, pp. 977–991, May 2024, doi: 10.1038/s41551-024-01207-5.

[54] G. K. Anumanchipalli, J. Chartier, and E. F. Chang, "Speech synthesis from neural decoding of spoken sentences," *Nature*, vol. 568, no. 7753, pp. 493–498, Apr. 2019, doi: 10.1038/s41586-019-1119-1.

**Mohamed Baha Ben Ticha** received the M.Sc. degree in computer science from the Higher School of Communications of Tunis, Tunisia, in 2019, and the Ph.D. degree in biomedical engineering from Université Grenoble Alpes, France, in 2025. He is currently a postdoctoral researcher at the Grenoble Institute of Neurosciences, working on speech brain–computer interfaces. His research interests include machine learning, representation learning, and domain adaptation, with applications to real-time decoding of neural signals into speech acoustics.

**Xingchen Ran** received his PhD degree in biomedical engineering from the Zhejiang University, Hangzhou, China, in 2022. From 2023 to 2025, he worked at NeuroXess Ltd., China. Since 2026, he has been an associate professor on the tenure-track at Chongqing University, Chongqing, China. His research interests include brain-computer interface, biomedical signal processing, and machine learning.

**Guillaume Saldanha** received the Engineering degree in biomedical engineering from the Grenoble INP - Phelma Engineering School of Grenoble, France, in 2021. He is currently a PhD student at the Grenoble Institute of Neurosciences, working on developing a real-time brain computer interface for speech communication. His research interests include brain computer interfaces, machine learning, signals processing, and neurosciences.

**Gaël Le Godais** received a master's degree in computer science from ENSIMAG, Grenoble in 2016, and a Ph.D. degree from Université Grenoble Alpes in 2022. He was a Postdoctoral Researcher at the Grenoble Institut des Neurosciences, where his research focused on decoding speech from brain activity. He is currently an independent composer and developer.

**Philémon Roussel** received M.S.E. degrees from École Centrale de Lyon and Keio University in 2017, and a Ph.D. in biomedical engineering from Université Grenoble Alpes in 2021. His research addresses fundamental and applied questions in neuroscience through computational analysis.

**Marc Aubert** received his engineering degree in computer science from ENSIMAG, Grenoble, in 2012. He worked as software developer for 3 years in private companies before joining the speech BCI project at the University Grenoble Alpes where he contributed to the development of software modules for neural data processing. He currently holds a position at Eugen Systems in the field of video games.

**Amina Fontanell** received her master's degree in biochemistry in 1995 and PhD in 2003 at Joseph Fourier University. She completed two years of postdoctoral research at the CEA in Grenoble. Since 2011, she has been a clinical research associate at Grenoble University Hospital, where she is currently responsible for regulatory aspects of clinical trials.

**Thomas Costecalde** obtained his PhD in Neuroscience from the University of Grenoble in 2012 in the field of BCI. He continued his work for several years in preclinical studies and also during a clinical study on BCI at Clinatec. He is currently in charge of the research program on the effects of thermobiomodulation at the CEA's SRBN laboratory.

**Lucas Struber** is a researcher at CEA-LETI's Division for Innovative Health Technologies in Grenoble, focusing on clinical uses of implanted ECoG-based brain–computer interfaces. He received an engineering MS in signal processing in 2013 and a PhD in motion analysis for rehabilitation in 2016, both from University Grenoble-Alpes. He subsequently shifted toward behavioral neuroscience and joined CEA Clinatec's BCI program in 2021 to combine neuroscience with movement restoration in advanced neurotechnology. He oversees patient BCI training, data processing to evaluate performance and progress, and the development of new clinical trials intended to broaden WIMAGINE clinical indications and future applications.

**Serpil Karakas** received the Engineering degree in microelectronics and automation from Polytech Montpellier, France, and the M.Sc. degree in Automatic Systems and Microelectronics from Université Montpellier II, France, in 2002. She is currently a Research Engineer at CEA, where she is responsible for the software platform within BCI/BSI projects. Her research interests include software development and regulatory documentation for brain–computer interface medical device applications.

**Shaomin Zhang** (IEEE Member) received his BSc and PhD in Biomedical Engineering from Zhejiang University in 2002 and 2007 respectively. He is currently a professor at the Qiushi Academy for Advanced Study at Zhejiang University in China. From 2004 to 2006, he was a research assistant under Prof. Raymond Tong at the Hong Kong Polytechnic University. From 2012 to 2014, he was a visiting scholar with Prof. John Donoghue at the Brown Institute for Brain Science at Brown University in the USA. His current research interests include neuroprosthetics, neurorehabilitation, invasive brain–machine interfaces and non-invasive neuromodulation.

**Philippe Kahane**, MD (1993), PhD (2003), is a Neurologist and University Professor of Physiology at Grenoble Alpes University and Hospital since 2007. He is recognized as an international expert in epilepsy surgery and intracranial EEG recordings, and is heading the Clinical Neuroscience Axis at Grenoble Alpes University Hospital.

**Guillaume Charvet** is Head of the Neurotechnology Biomedical Research Unit at CEA Leti CLINATEC® in Grenoble, France, where he leads the design and integration of medical devices addressing clinical needs. His research focuses on neuroengineering for neuroprosthetics and neuromodulation systems. Since 2008, he has been involved in developing the WIMAGINE® brain–computer interface technology, based on a fully implantable device enabling wireless brain signal recording and neural decoding. This technology is evaluated in clinical proof-of-concept studies for movement restoration in individuals with severe motor disabilities. He coordinates and contributes to several national and international institutional research projects within collaborative frameworks.

**Stephan Chabardès**, MD, PhD, is Neurosurgeon at Grenoble University Hospital and Professor of Neurosurgery at Grenoble Alpes University since 2009. Trained in Grenoble under Professor Alim-Louis Benabid, he completed a research fellowship at the Cleveland Clinic. His research has focused on neuromodulation and innovative neurotechnologies, with a particular emphasis on advancing surgical approaches for drug-resistant epilepsy. He has been heading the Department of Neurosurgery since 2021 and is Scientific Director of Clinatec since 2025. Author of over 215 peer-reviewed publications, he is an active leader in international societies for stereotactic and functional neurosurgery.

**Blaise Yvert** is senior researcher at INSERM. He received his Engineering degrees from Ecole Centrale de Lyon and Cornell University in 1993, his PhD in Biomedical Engineering from INSA Lyon in 1996, and his habilitation (HDR) in 2008. In 2012- 2013 he became appointed Fulbright Visiting Scholar at the Brown Institute for Brain Sciences directed by John P. Donoghue in the field of human brain-computer interfaces (BCIs). Since 2013, his research focusses on the development of new cortical interfaces and BCI systems for speech rehabilitation.